\documentclass[a4paper, 10pt]{article}

\usepackage{epsfig}
\usepackage{amsmath}

\newcommand{\bel}{\mathrm{bel}}
\newcommand{\pl}{\mathrm{pl}}
\newcommand{\betP}{\mathrm{betP}}
\newcommand{\Bel}{\mathrm{Bel}}
\newcommand{\Pl}{\mathrm{Pl}}
\newcommand{\GPT}{\mathrm{GPT}}
\newcommand{\eR}{\mbox{I\hspace{-.15em}R}}
\newcommand{\ind}{\mbox{1\hspace{-.25em}l}}

\newcommand{\PCRmo}{\mathrm{PCR6}}
\newcommand{\PCRdsm}{\mathrm{PCR5}}
\newcommand{\Pmof}{\mathrm{PCRf}}
\newcommand{\Pmog}{\mathrm{PCRg}}
\newcommand{\conj}{\mathrm{c}}
\newcommand{\DP}{\mathrm{DP}}

\title{Generalized proportional conflict redistribution rule applied to Sonar imagery and Radar targets classification}
\author{Arnaud Martin and Christophe Osswald}

\begin{document}
\maketitle

\begin{abstract}
In this chapter, we present two applications in information fusion in order to evaluate the generalized proportional conflict redistribution rule presented in the chapter \cite{Martin06a}. Most of the time the combination rules are evaluated only on simple examples. We study here different combination rules and compare them in terms of decision on real data. Indeed, in real applications, we need a reliable decision and it is the final results that matter. Two applications are presented here: a fusion of human experts opinions on the kind of underwater sediments depict on sonar image and a classifier fusion for radar targets recognition.

{\bf Keywords:} Experts fusion, classification, DST, DSmT, generalized PCR, Sonar, Radar.
\end{abstract}

\section{Introduction}
We have presented and discussed on some combination rules in the chapter \cite{Martin06a}. Our study was essentially on the redistribution of conflict rules. We have proposed a new proportional conflict redistribution rule. We have seen that the decision can be different following the rule. Most of the time the combination rules are evaluated only on simple examples. In this chapter, we study different combination rules and compare them in terms of decision on real data. Indeed, in real applications, we need a reliable decision and it is the final results that matter. Hence, for a given application, the best combination rule is the rule given the best results. For the decision step, different functions such as credibility, plausibility and pignistic probability \cite{Shafer76, Smets90, Dezert04} are usually used. 

In this chapter, we present the advantages of the DSmT for the modelization of real applications and also for the combination step. First, the principles of the DST and DSmT are recalled. We present the formalization of the belief function models, different rules of combination and decision. One the combination rule (PCR5) proposed by \cite{Smarandache05} for two experts is mathematically one of the best for the proportional redistribution of the conflict applicable in the context of the DST and the DSmT. We compare here an extension of this rule for more experts, the PCR6 rule presented in the chapter \cite{Martin06a}. 

Two applications are presented here: a fusion of human experts opinions on the kind of underwater sediments depict on sonar image and a classifier fusion for radar targets recognition.

The first application relates the seabed characterization, for instance in order to help the navigation of Autonomous Underwater Vehicles or provide data to sedimentologists. The sonar images are obtained with many imperfections due to instrumentations measuring a huge number of physical data (geometry of the device, coordinates of the ship, movements of the sonar, etc.). In this kind of applications, the reality is unknown. If human experts have to classify sonar images they can not provide with certainty the kind of sediment on the image. Thus, for instance, in order to train an automatic classification algorithm, we must take into account this difference and the uncertainty of each expert. We propose in this chapter how to solve this human expert fusion.

The second application allows to really compare the combination rules. We present an application of classifier fusion in order to extract the information for the automatic target recognition. The real data are provided by measures in the anechoic chamber of ENSIETA (Brest, France) obtained illuminating 10 scale reduced (1:48) targets of planes. Hence, all the experimentations are controlled and the reality is known. The results of the fusion of three classifiers are studied in terms of good-classification rates.

This chapter is organized as follow: In the first section, we recall combination rules presented in the chapter \cite{Martin06a} and we compare in this chapter. The section \ref{sonar} proposes a mean to fuse human expert's opinions in uncertain environments such as the underwater milieu. This environment is described with sonar images the most appropriate in such environment. The last section presents the results of classifiers fusion in an application of radar targets recognition.

\section{Backgrounds on combination rules}
We recall here the combination rules presented and discussed in the chapter \cite{Martin06a} and compared on two real applications in the forwards sections. For more details on the theory bases see the chapter \cite{Martin06a}.

In the context of the DST, the non-normalized conjunctive rule is one of the most used rule and is given by \cite{Smets90} for all $X \in 2^\Theta$ by:
\begin{eqnarray}
\label{conjunctive}
m_\conj(X)=\sum_{Y_1 \cap ... \cap Y_M = X} \prod_{j=1}^M m_j(Y_j),
\end{eqnarray}
where $Y_j \in 2^\Theta$ is the response of the expert $j$, and $m_j(Y_j)$ the associated basic belief assignments.

In this chapter, we focus on rules where the conflict is redistributed. With the rule given in the Dubois and Prade rule \cite{Dubois88}, a mixed conjunctive and disjunctive rule, the conflict is redistributed on partial ignorance. This rule is given for all $X \in 2^\Theta$, $X\neq \emptyset$ by: 
\begin{eqnarray}
\label{DP}
m_\DP(X)=\sum_{Y_1 \cap ... \cap Y_M = X} \prod_{j=1}^M m_j(Y_j)+\sum_{
\begin{array}{c}
\scriptstyle Y_1 \cup ... \cup Y_M = X\\
\scriptstyle Y_1 \cap ... \cap Y_M = \emptyset \\
\end{array}} \prod_{j=1}^M m_j(Y_j),
\end{eqnarray}
where $Y_j \in 2^\Theta$ is the response of the expert $j$, and $m_j(Y_j)$ the associated basic belief assignments.

In the context of the DSmT, the non-normalized conjunctive rule can be used for all $X \in D^\Theta$ and $Y \in D^\Theta$. The mixed rule given by the equation (\ref{DP}) has been rewrite in \cite{Smarandache04}, and recalled DSmH, for all $X \in D^\Theta$, $X\not\equiv \emptyset$ \footnote{The notation $X\not\equiv \emptyset$ means that $X \neq \emptyset$ and following the chosen model in $D^\Theta$, $X$ is not one of the element of $D^\Theta$ defined as $\emptyset$. For example, if $\Theta=\{A, B, C\}$, we can define a model for which the expert can provide a mass on $A\cap B$ and not on $A \cap C$, so $A\cap B\neq \emptyset$ and $A\cap B=\emptyset$} by:
\begin{equation}
  \label{DSmH}
  \begin{array}{c}
    m_H(X)=\displaystyle \sum_{Y_1 \cap ... \cap Y_M = X} \prod_{j=1}^M m_j(Y_j)+\displaystyle \sum_{
      \begin{array}{c}
	\scriptstyle Y_1 \cup ... \cup Y_M = X\\
	\scriptstyle Y_1 \cap ... \cap Y_M \equiv \emptyset \\
    \end{array}} 
    \prod_{j=1}^M m_j(Y_j)+\\ 
    \displaystyle \sum_{
      \begin{array}{c}
	\scriptstyle \left\{u(Y_1) \cup ... \cup u(Y_M) = X\right\} \\ 
	\scriptstyle Y_1, ..., Y_M \equiv \emptyset \\
    \end{array}} \displaystyle \!\!\!\!\prod_{j=1}^M m_j(Y_j)+
    \!\!\!\!\!\!\!\!\displaystyle \sum_{
      \begin{array}{c}
	\scriptstyle \left\{ u(Y_1) \cup ... \cup u(Y_M) \equiv \emptyset \, \mbox{and} \, X=\Theta\right\} \\
	\scriptstyle Y_1, ..., Y_M \equiv \emptyset \\
    \end{array}} \displaystyle \prod_{j=1}^M m_j(Y_j),
  \end{array}
\end{equation}
where $Y_j \in D^\Theta$ is the response of the expert $j$, $m_j(Y_j)$ the associated basic belief assignments, and $u(Y)$ is the function giving the union that compose $Y$ \cite{Smarandache04chap4}. For example if $Y=(A\cap B) \cup (A \cap C)$, $u(Y)=A \cup B \cup C$.

If we want to take the decision only on the elements in $\Theta$, some rules propose to redistribute the conflict proportionally on these elements. The most accomplished is the PCR5 given in \cite{Smarandache05}. The equation for $M$ experts, for $X\in D^\Theta$, $X\not\equiv \emptyset$ is given in \cite{Dezert06} by:
\begin{eqnarray}
  \label{GenePCR_DS}
  & & \!\!\!\!\!\!\!\!\!\!\!\!\!\!\! \nonumber \displaystyle \!\!\!m_\PCRdsm(X) = m_\conj(X) + \\
  & & \!\!\!\!\!\!\!\!\!\!\!\!\!\!\! \sum_{i=1}^M
  m_i(X) \!\!\!\!\!\!\!\!\!\!\!\!\!\!\!\!\!\!\!\!\!\!\!\!\! \sum_{\begin{array}{c}
      \scriptstyle (Y_{\sigma_i(1)},...,Y_{\sigma_i(M\!-\!1)})\in
      (D^\Theta)^{M\!-\!1} \\
      \scriptstyle {\displaystyle \mathop{\cap}_{\scriptscriptstyle
	  k=1}^{\scriptscriptstyle M\!-\!1}} Y_{\sigma_i(k)} \cap X
      \equiv \emptyset
  \end{array}}
  \!\!\!\!\!\!\!\!\!\!\!\!\!\!\!\frac{
    \displaystyle \Bigg(\prod_{j=1}^{M\!-\!1} m_{\sigma_i(j)}(Y_{\sigma_i(j)})\ind_{j>i}\Bigg)
    \!\!\!\prod_{Y_{\sigma_i(j)}=X} \!\!\!\!\! m_{\sigma_i(j)}(Y_{\sigma_i(j)})
  }{
    \displaystyle \!\!\!\!\!\!\!\!\!\!\!\!\sum_{~~~~~\renewcommand{\arraystretch}{1.8}\begin{array}{c}
	\scriptstyle Z\in\{X, Y_{\sigma_i(1)}, \ldots, Y_{\sigma_i(M\!-\!1)}\}
    \end{array}}
    \!\!\!\!\!\!\!\!\!\!\!\!\!\!\!\!\!\!\!\!\!\!\!\!\!\!\!\!\!\!\!\!\!\prod_{Y_{\sigma_i(j)}=Z}^{\:}\!\!\!\!\!\!\!
    \big(m_{\sigma_i(j)}(Y_{\sigma_i(j)}).T({\scriptstyle
    X\!=\!Z,m_i(X)})\big)},
\end{eqnarray}
where $\sigma_i$ counts from 1 to $M$ avoiding $i$:
\begin{eqnarray}
\label{sigma}
\left\{
\begin{array}{ll}
\sigma_i(j)=j &\mbox{if~} j<i,\\
\sigma_i(j)=j+1 &\mbox{if~} j\geq i,\\
\end{array}
\right.
\end{eqnarray}
and:
\begin{eqnarray}
\label{T}
\left\{
\begin{array}{ll}
T(B,x)=x &\mbox{if $B$ is true},\\
T(B,x)=1 &\mbox{if $B$ is false},\\
\end{array}
\right.
\end{eqnarray}

We have proposed another proportional conflict redistribution PCR6 rule in the chapter \cite{Martin06a}, for $M$ experts, for $X\in D^\Theta$, $X\neq \emptyset$:
\begin{eqnarray}
\label{GeneDSmTcombination}
  \displaystyle m_\PCRmo(X) & = & \displaystyle m_\conj(X) + \\
  \nonumber & & \!\!\!\!\!\!\!\!\!\!\sum_{i=1}^M m_i(X)^2
  \!\!\!\!\!\!\!\!\!\!\!\!\!\!\!\!\!\!\!\! \sum_{\begin{array}{c}
      \scriptstyle {\displaystyle \mathop{\cap}_{k=1}^{M\!-\!1}} Y_{\sigma_i(k)} \cap X \equiv \emptyset \\
      \scriptstyle (Y_{\sigma_i(1)},...,Y_{\sigma_i(M\!-\!1)})\in (D^\Theta)^{M\!-\!1}
  \end{array}}
  \!\!\!\!\!
  \left(\!\!\frac{\displaystyle \prod_{j=1}^{M\!-\!1} m_{\sigma_i(j)}(Y_{\sigma_i(j)})}
       {\displaystyle m_i(X) \!+\! \sum_{j=1}^{M\!-\!1}
  m_{\sigma_i(j)}(Y_{\sigma_i(j)})}\!\!\right)\!\!,
\end{eqnarray}
where $\sigma$ is defined like in (\ref{sigma}). \\

$m_i(X)+\displaystyle \sum_{j=1}^{M-1} m_{\sigma_i(j)}(Y_{\sigma_i(j)}) \neq 0$, $m_c$ is the conjunctive consensus rule given by the equation (\ref{conjunctive}). The PCR6 and PCR5 rules are exactly the same for in the case of 2 experts.

We have also proposed two more generalized rules given by:

\begin{eqnarray}
  \label{GenePCR_f}
   \displaystyle m_{\Pmof}(X) & = & \displaystyle m_\conj(X) + \\
  \nonumber & &
  \!\!\!\!\!\!\!\!\!\!\!\!\!\!\!\!\!\!\!\!\!\!\!\!\!\!\!\!\!\!\!\!\!\!\!\!\!\!\!\!\!\!\!\!\!\!\!\!
  \sum_{i=1}^M  m_i(X)f(m_i(X)) 
  \!\!\!\!\!\!\!\!\!\!\!\!\!\!\!\!\!\!\!\!\!\!\!
  \displaystyle \sum_{\begin{array}{c}
	\scriptstyle {\displaystyle \mathop{\cap}_{k=1}^{M\!-\!1}} Y_{\sigma_i(k)} \cap X \equiv \emptyset \\
	\scriptstyle (Y_{\sigma_i(1)},...,Y_{\sigma_i(M\!-\!1)})\in (D^\Theta)^{M\!-\!1}
    \end{array}}
    \!\!\!\!\!\!
    \left(\!\!\frac{\displaystyle \prod_{j=1}^{M\!-\!1} m_{\sigma_i(j)}(Y_{\sigma_i(j)})}
	 {\displaystyle f(m_i(X)) \!+\! \sum_{j=1}^{M\!-\!1} f(m_{\sigma_i(j)}(Y_{\sigma_i(j)}))}\!\!\right)\!\!,
\end{eqnarray}
with the same notations that in the equation (\ref{GeneDSmTcombination}), and $f$ an increasing function defined by the mapping of $[0,1]$ onto $\eR^+$. 

The second generalized rule is given by:
\begin{eqnarray}
  \label{GenePCR_g}
  \begin{array}{l}
    \displaystyle m_{\Pmog}(X) =  m_\conj(X) + 
    \displaystyle \sum_{i=1}^M \!\!\!\!\!\!\!\!\!\!\!\!
    \sum_{\begin{array}{c}
	\scriptstyle {\displaystyle \mathop{\cap}_{\scriptscriptstyle
	    k=1}^{\scriptscriptstyle M\!-\!1}} Y_{\sigma_i(k)} \cap X
	\equiv \emptyset\\ 
	\scriptstyle (Y_{\sigma_i(1)},...,Y_{\sigma_i(M\!-\!1)})\in (D^\Theta)^{M\!-\!1}
    \end{array}}\\ \\
     m_i(X) \frac{\displaystyle
      \Bigg(\!\prod_{j=1}^{M\!-\!1}\!\!m_{\sigma_i(j)}(Y_{\sigma_i(j)})\!\!\Bigg)
      \!\!\Bigg(\!\!\!\!\!\!\!\prod_{~~~Y_{\sigma_i(j)}=X}\!\!\!\!\!\!\!\!\!\!\!\ind_{j>i}\!\!\Bigg)
      g\Bigg(\!\!m_i(X)\!\!+\!\!\!\!\!\!\!\!\sum_{Y_{\sigma_i(j)}=X}\displaystyle \!\!\!\!\!\!\!\!m_{\sigma_i(j)}(Y_{\sigma_i(j)})\!\!\!\Bigg)}
	{\displaystyle \sum_{
	    \scriptstyle Z\in\{X, Y_{\sigma_i(1)}, \ldots, Y_{\sigma_i(M\!-\!1)}\}}
	  \!\!\!g\!\!\left(\!\sum_{Y_{\sigma_i(j)}=Z}\!\!\!\!\!\displaystyle m_{\sigma_i(j)}(Y_{\sigma_i(j)})+m_i(X)\ind_{\scriptscriptstyle X=Z}\!\!\!\right)},
  \end{array}
\end{eqnarray}

with the same notations that in the equation (\ref{GeneDSmTcombination}), and $g$ an increasing function defined by the mapping of $[0,1]$ onto $\eR^+$.

In this chapter, we choose $f(x)=g(x)=x^\alpha$, with $\alpha \in \eR^+$.

\section{Experts fusion in Sonar imagery}
\label{sonar}

\begin{figure}[htb]
\includegraphics[height=4.7cm]{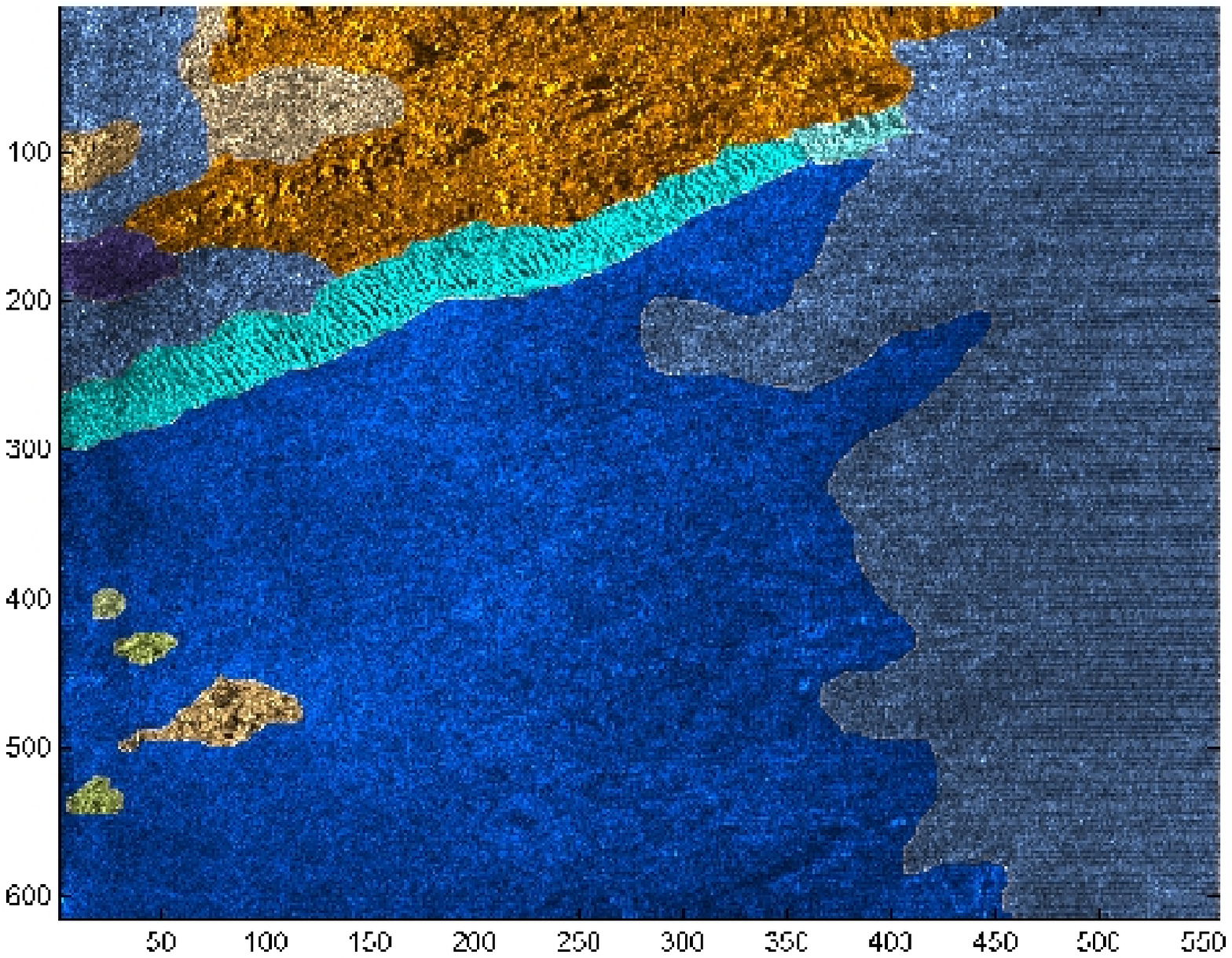}
\includegraphics[height=4.7cm]{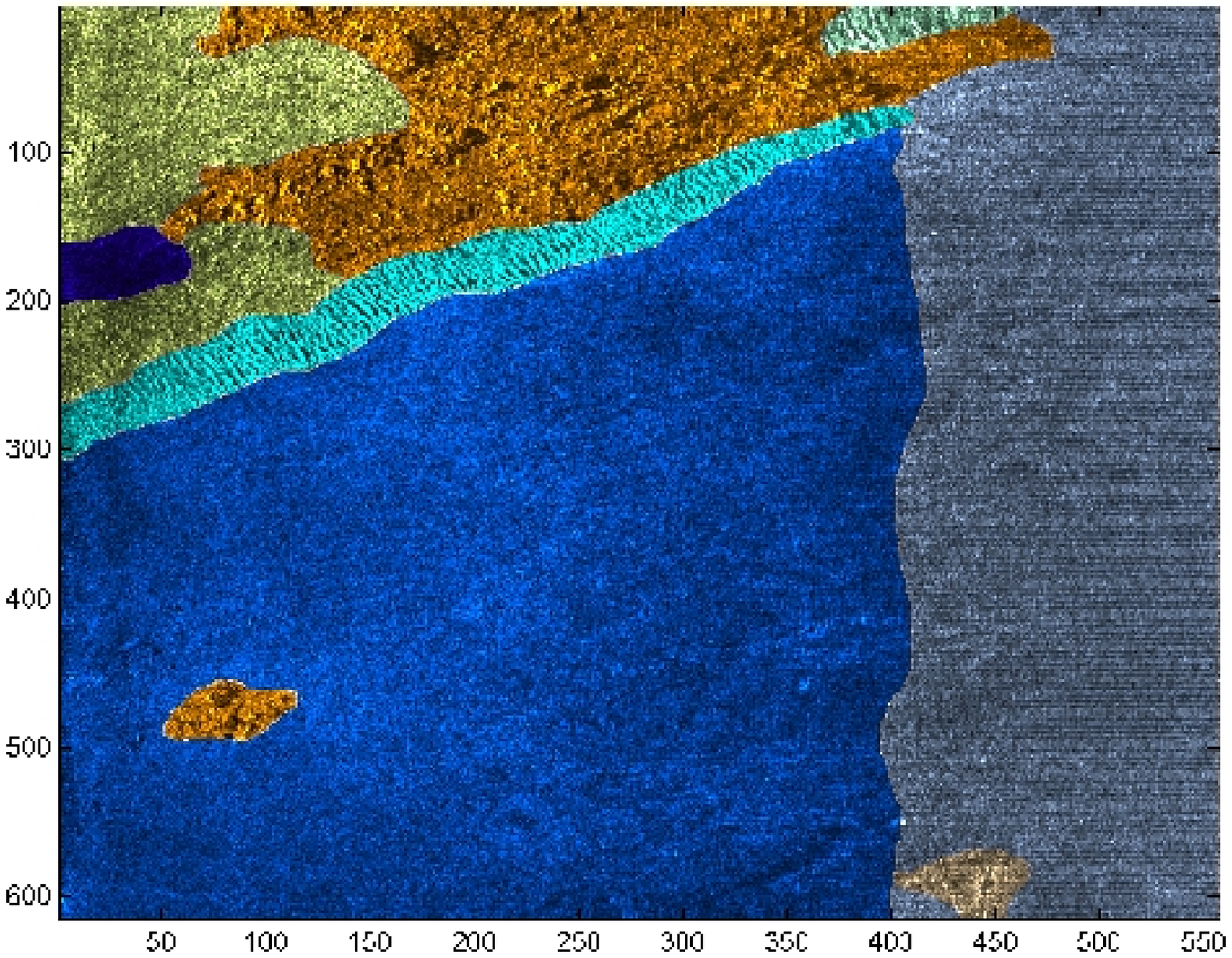}
\vspace{-0.5cm}
\caption{Segmentation given by two experts.}
\label{expert}
\end{figure}

Seabed characterization serves many useful purposes, {\it e.g.} help the navigation of Autonomous Underwater Vehicles or provide data to sedimentologists. In such sonar applications, seabed images are obtained with many imperfections \cite{Martin05}. Indeed, in order to build images, a huge number of physical data (geometry of the device, coordinates of the ship, movements of the sonar, etc.) has to be taken into account, but these data are polluted with a large amount of noises caused by instrumentations. In addition, there are some interferences due to the signal traveling on multiple paths (reflection on the bottom or surface), due to speckle, and due to fauna and flora. Therefore, sonar images have a lot of imperfections such as imprecision and uncertainty; thus sediment classification on sonar images is a difficult problem. In this kind of applications, the reality is unknown and different experts can propose different classifications of the image. Figure \ref{expert} exhibits the differences between the interpretation and the certainty of two sonar experts trying to differentiate the type of sediment (rock, cobbles, sand, ripple, silt) or shadow when the information is invisible. Each color corresponds to a kind of sediment and the associated certainty of the expert for this sediment expressed in term of sure, moderately sure and not sure. Thus, in order to train an automatic classification algorithm, we must take into account this difference and the uncertainty of each expert. Indeed, image classification is generally done on a local part of the image (pixel, or most of the time on small tiles of {\it e.g.} 16$\times$16 or 32$\times$32 pixels). For example, how a tile of rock labeled as {\it not sure} must be taken into account in the learning step of the classifier and how to take into account this tile if another expert says that it is sand? Another problem is: how should we consider a tile with more than one sediment?

In this case, the space of discernment $\Theta$ represents the different kind of sediments on sonar images, such as rock, sand, silt, cobble, ripple or shadow (that means no sediment information). The experts give their perception and belief according to their certainty. For instance, the expert can be moderately sure of his choice when he labels one part of the image as belonging to a certain class, and be totally doubtful on another part of the image. Moreover, on a considered tile, more than one sediment can be present. 

Consequently we have to take into account all these aspects of the applications. In order to simplify, we consider only two classes in the following: the rock referred as $A$, and the sand, referred as $B$. The proposed models can be easily extended, but their study is easier to understand with only two classes.

Hence, on certain tiles, $A$ and $B$ can be present for one or more experts. The belief functions have to take into account the certainty given by the experts (referred respectively as $c_A$ and $c_B$, two numbers in $[0,1]$) as well as the proportion of the kind of sediment in the tile $X$ (referred as $p_A$ and $p_B$, also two numbers in $[0,1]$). We have two interpretations of ``the expert believes $A$'': it can mean that the expert thinks that there is $A$ on $X$ and not $B$, or it can mean that the expert thinks that there is $A$ on $X$ and it can also have $B$ but he does not say anything about it. The first interpretation yields that hypotheses $A$ and $B$ are exclusive and with the second they are not exclusive. We only study the first case: $A$ and $B$ are exclusive. But on the tile $X$, the expert can also provide $A$ and $B$, in this case the two propositions ``the expert believes $A$'' and ``the expert believes $A$ and $B$'' are not exclusive.

\subsection{Models}
We have proposed five models and studied these models for the fusion
of two experts \cite{Martin06}. We present here the three last models for two experts and
two classes. In this case the conjunctive rule (\ref{conjunctive}), the mixed rule (\ref{DP}) and the
DSmH (\ref{DSmH}) are similar. We give the obtained results on a real database for the fusion of three experts in
sonar.  

\begin{figure}[htb]
%\vspace{-0.5cm}
\begin{center}
\includegraphics[height=5cm]{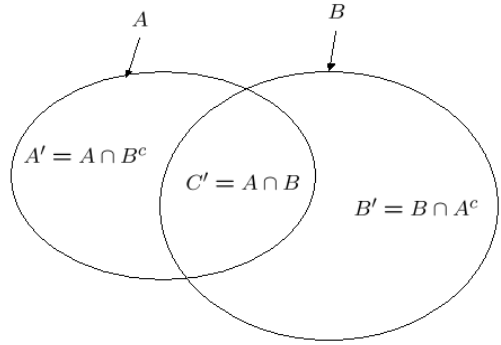}
\end{center}
\vspace{-0.5cm}
\caption{Notation of the intersection of two classes $A$ and $B$.}
\label{AetB}
\end{figure}

\paragraph{Model $M_3$}
In our application, $A$, $B$ and $C$ cannot be considered exclusive on $X$. In order to propose a model following the DST, we have to study exclusive classes only. Hence, in our application, we can consider a space of discernment of three exclusive classes $\Theta=\{A \cap B^c, B\cap A^c, A \cap B\}=\{A', B', C'\}$, following the notations given on the figure \ref{AetB}.

Hence, we can propose a new model $M_3$ given by:
\begin{eqnarray}
\label{M3}
\begin{array}{l}
\mbox{if the expert says $A$:} \\
  \left\{
  \begin{array}{l}
  m(A'\cup C')=c_A, \\
  m(A' \cup B' \cup C')=1-c_A,
  \end{array}
  \right. \\
  \\
\mbox{if the expert says $B$:}\\
	\left\{
  \begin{array}{l}
  m(B'\cup C')=c_B, \\
  m(A' \cup B' \cup C')=1-c_B,
  \end{array}
  \right.  \\
  \\
\mbox{if the expert says $C'$:}\\
  \left\{
  \begin{array}{l}
  m(C')=p_A . c_A + p_B . c_B, \\
  m(A' \cup B' \cup C')=1-(p_A . c_A + p_B . c_B).
  \end{array}
  \right.
\end{array}
\end{eqnarray}
Note that $A' \cup B' \cup C'=A\cup B$. On our numerical example we obtain:
\begin{eqnarray*}
  \begin{array}{|c|c|c|c|c|}
  \hline
   & A'\cup C' & B' \cup C' & C' & A' \cup B' \cup C'\\
  \hline
   m_1 & 0.6 & 0 & 0 & 0.4 \\
  \hline
   m_2 & 0 & 0 & 0.5 & 0.5 \\
  \hline  
  \end{array}
\end{eqnarray*}

Hence, the conjunctive rule, the credibility, the plausibility and the pignistic probability are given by:
\begin{eqnarray*}
  \begin{array}{|c|c|c|c|c|}
  \hline
  element & m_c & \bel & \pl & \betP \\
  \hline
\emptyset & 0 & 0 & 0 & - \\
  \hline
A'=A \cap B^c &0 & 0 & 0.5 & 0.2167 \\
  \hline
B'=B\cap A^c & 0 & 0& 0.2 & 0.0667 \\
  \hline
A'\cup B'=(A \cap B^c)\cup (B\cap A^c)& 0& 0 & 0.5 & 0.2833 \\
  \hline
C'=A \cap B & 0.5 & 0.5 & 1 & 0.7167\\
  \hline
A' \cup C'=A & 0.3 & 0.8 & 1 & 0.9333 \\
  \hline
B'\cup C'=B & 0 & 0.5  & 1 & 0.7833 \\
  \hline
A'\cup B' \cup C'=A\cup B & 0.2 & 1 & 1 & 1 \\
  \hline
  \end{array}
\end{eqnarray*}
where
\begin{eqnarray}
m_c(C')=m_c(A\cap B)=0.2+0.3=0.5.
\end{eqnarray}

In this example, with this model $M_3$ the decision will be $A$ with the maximum of the pignistic probability. But the decision could {\it a priori} be taken also on $C'=A\cap B$ because $m_c(C')$ is the highest. We have seen that if we want to take the decision on $A\cap B$, we must considered the maximum of the masses because of inclusion relations of the credibility, plausibility and pignistic probability.

\paragraph{Model $M_4$}
In the context of the DSmT, we can write $C=A\cap B$ and easily propose a fourth model $M_4$, without any consideration on the exclusivity of the classes, given by:

\begin{eqnarray}
\label{M4}
\begin{array}{l}
\mbox{if the expert says $A$:}\\
  \left\{
  \begin{array}{l}
  m(A)=c_A, \\
  m(A \cup B)=1-c_A,
  \end{array}
  \right. \\
  \\
\mbox{if the expert says $B$:}\\
	\left\{
  \begin{array}{l}
  m(B)=c_B, \\
  m(A \cup B)=1-c_B,
  \end{array}

  \right.  \\
  \\
\mbox{if the expert says $A\cap B$:}\\
  \left\{
  \begin{array}{l}
  m(A \cap B)=p_A . c_A + p_B . c_B, \\
  m(A \cup B)=1-(p_A . c_A + p_B . c_B).
  \end{array}
  \right.
\end{array}
\end{eqnarray}
This last model $M_4$ allows to represent our problem without adding an artificial class $C$. Thus, the model $M_4$ based on the DSmT gives:

\begin{eqnarray*}
  \begin{array}{|c|c|c|c|c|}
  \hline
   & A & B & A \cap B & A \cup B \\
  \hline
   m_1 & 0.6 & 0 & 0 & 0.4 \\
  \hline
   m_2 & 0 & 0 & 0.5 & 0.5 \\
  \hline  
  \end{array}
\end{eqnarray*}

The obtained mass $m_c$ with the conjunctive  yields:
  
\begin{eqnarray}
\label{M4conjunctive}
  \begin{array}{l}
  m_c(A)=0.30, \\
  m_c(B)=0, \\
  m_c(A \cap B)=m_1(A)m_2(A\cap B)+  m_1(A\cup B)m_2(A\cap B)\\
  \quad \quad \quad \quad \quad=0.30+0.20=0.5,\\
  m_c(A\cup B)=0.20. \\
  \end{array}
\end{eqnarray}

These results are exactly similar to the model $M_3$. These two models do not present ambiguity and show that the mass on $A\cap B$ (rock and sand) is the highest. 

The  generalized credibility, the  generalized plausibility and the generalized pignistic probability are given by:
\begin{eqnarray*}
  \begin{array}{|c|c|c|c|c|}
  \hline
  element  & m_c & \Bel & \Pl & \GPT \\
  \hline
\emptyset & 0 & 0  & 0 & - \\
  \hline
A & 0.3&  0.8 & 1 & 0.9333 \\
  \hline
B & 0 & 0.5 & 0.7& 0.7833 \\
  \hline
A\cap B& 0.5& 0.5 & 1 & 0.7167 \\
  \hline
A\cup B & 0.2 & 1  & 1 & 1 \\
  \hline
  \end{array}
\end{eqnarray*}

Like the model $M_3$, on this example, the decision will be $A$ with the maximum of pignistic probability criteria. But here also the maximum of $m_c$ is reached for $A\cap B=C'$.

If we want to consider only the kind of possible sediments $A$ and $B$ and do not allow their conjunction, we can use a proportional conflict redistribution rule such as the PCR rule:
\begin{eqnarray}
\label{M4PCR}
  \begin{array}{l}
  m_{PCR}(A)=0.30+0.5=0.8, \\
  m_{PCR}(B)=0, \\
  m_{PCR}(A\cup B)=0.20. \\
  \end{array}
\end{eqnarray}

The credibility, the plausibility and the pignistic probability are given by:
\begin{eqnarray*}
  \begin{array}{|c|c|c|c|c|}
  \hline
  element  & m_{PCR} & \bel & \pl & \betP \\
  \hline
\emptyset & 0 & 0  & 0 & - \\
  \hline
A & 0.8&  0.8 & 1 & 0.9 \\
  \hline
B & 0 & 0& 0.2& 0.1 \\
  \hline
A\cup B & 0.2 & 1  & 1 & 1 \\
  \hline
  \end{array}
\end{eqnarray*}
On this numerical example, the decision will be the same than the conjunctive rule, here the maximum of pignistic probability is reached for $A$ (rock). In the next section we see that is not always the case.

\paragraph{Model $M_5$}
Another model $M_5$ which can be used in both the DST and the DSmT is given considering only one belief function according to the proportion by:
\begin{eqnarray}
\label{M5}
\begin{array}{l}
  \left\{
  \begin{array}{l}
  m(A)=p_A.c_A, \\
  m(B)=p_B.c_B,\\
  m(A \cup B)=1-(p_A . c_A + p_B . c_B).
  \end{array}
  \right. \\
\end{array}
\end{eqnarray}
If for one expert, the tile contains only $A$, $p_A=1$, and $m(B)=0$. If for another expert, the tile contains $A$ and $B$, we take into account the certainty and proportion of the two sediments but not only on one focal element. Consequently, we have simply:
\begin{eqnarray*}
  \begin{array}{|c|c|c|c|}
  \hline
   & A & B & A \cup B \\
  \hline
   m_1 & 0.6 & 0  & 0.4 \\
  \hline
   m_2 & 0.3& 0.2 & 0.5 \\
  \hline  
  \end{array}
\end{eqnarray*}

In the DST context, the conjunctive rule, the credibility, the plausibility and the pignistic probability are given by:
\begin{eqnarray*}
  \begin{array}{|c|c|c|c|c|}
  \hline
  element  & m_c & \bel & \pl & \betP \\
  \hline
\emptyset & 0.12 & 0  & 0 & - \\
  \hline
A & 0.6&  0.6 & 0.8 & 0.7955 \\
  \hline
B & 0.08 & 0.08& 0.28& 0.2045 \\
  \hline
A\cup B & 0.2 & 0.88  & 0.88 & 1 \\
  \hline
  \end{array}
\end{eqnarray*}
In this case we do not have the plausibility to decide on $A\cap B$, because the conflict is on $\emptyset$.

In the DSmT context, the conjunctive rule, the generalized credibility, the generalized plausibility and the generalized pignistic probability are given by:
\begin{eqnarray*}
  \begin{array}{|c|c|c|c|c|}
  \hline
  element  & m_c & \Bel & \Pl & \GPT \\
  \hline
\emptyset & 0 & 0  & 0 & - \\
  \hline
A & 0.6&  0.72 & 0.92 & 0.8933 \\
  \hline
B & 0.08 & 0.2& 0.4& 0.6333 \\
  \hline
A\cap B& 0.12& 0.12 & 1 & 0.5267 \\
  \hline
A\cup B & 0.2 & 1  & 1 & 1 \\
  \hline
  \end{array}
\end{eqnarray*}
The decision with the maximum of pignistic probability criteria is still $A$.

The PCR rule provides:
\begin{eqnarray*}
  \begin{array}{|c|c|c|c|c|}
  \hline
  element  & m_{PCR} & \bel & \pl & \betP \\
  \hline
\emptyset & 0 & 0  & 0 & - \\
  \hline
A & 0.69&  0.69 & 0.89 & 0.79 \\
  \hline
B & 0.11 & 0.11& 0.31& 0.21 \\
  \hline
A\cup B & 0.2 & 1  & 1 & 1 \\
  \hline
  \end{array}
\end{eqnarray*}
where
\begin{eqnarray*}
  \begin{array}{l}
  m_{PCR}(A)=0.60+0.09=0.69, \\
  m_{PCR}(B)=0.08+0.03=0.11. \\
  \end{array}
\end{eqnarray*}
With this model and example the PCR rule, the decision will be also $A$, and we do not have difference between the conjunctive rules in the DST and DSmT.

\subsection{Experimentation}
\label{illustration}

\paragraph{Database}
Our database contains 42 sonar images provided by the GESMA (Groupe d'Etudes Sous-Marines de l'Atlantique). These images were obtained with a Klein 5400 lateral sonar with a resolution of 20 to 30 cm in azimuth and 3 cm in range. The sea-bottom depth was between 15 m and 40 m.

\begin{figure}[htb]
\begin{center}
\includegraphics[height=5cm]{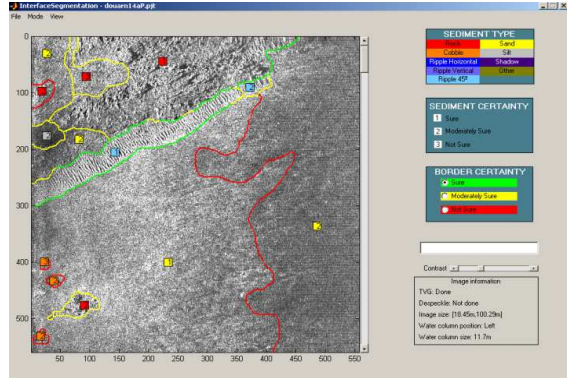}
\end{center}
\caption{Manual Segmentation Interface.}
\label{manual_seg}
\end{figure}

Three experts have manually segmented these images giving the kind of sediment (rock, cobble, sand, silt, ripple (horizontal, vertical or at 45 degrees)), shadow or other (typically ships) parts on images, helped by the manual segmentation interface presented in figure \ref{manual_seg}. All sediments are given with a certainty level (sure, moderately sure or not sure). Hence, each pixel of every image is labeled as being either a certain type of sediment or a shadow or other.

The three experts provide respectively, 30338, 31061, and 31173 homogeneous tiles, 8069, 7527, and 7539 tiles with two sediments, 575, 402, and 283 tiles with three sediments, 14, 7, and 2 tiles with four, and 1, 0, and 0 tile for five sediments, and 0 for more.

\paragraph{Results}
We note $A=$ rock, $B=$ cobble, $C=$ sand, $D=$ silt, $E=$ ripple, $F=$ shadow and $G=$ other, hence we have seven classes and $\Theta=\{A,B,C,D,E,F,G\}$. We applied the generalized model $M_5$ on tiles of size 32$\times$32 given by:
\begin{eqnarray}
\label{GeneralizedM5}
\begin{array}{l}
  \left\{
  \begin{array}{l}
  m(A)=p_{A1}.c_1+p_{A2}.c_2+p_{A3}.c_3, \, \mbox{for rock,}\\
  m(B)=p_{B1}.c_1+p_{B2}.c_2+p_{B3}.c_3, \, \mbox{for cobble,}\\
  m(C)=p_{C1}.c_1+p_{C2}.c_2+p_{C3}.c_3, \, \mbox{for ripple,}\\
  m(D)=p_{D1}.c_1+p_{D2}.c_2+p_{D3}.c_3, \, \mbox{for sand,}\\
  m(E)=p_{E1}.c_1+p_{E2}.c_2+p_{E3}.c_3, \, \mbox{for silt,}\\
  m(F)=p_{F1}.c_1+p_{F2}.c_2+p_{F3}.c_3, \, \mbox{for shadow,}\\
  m(G)=p_{G1}.c_1+p_{G2}.c_2+p_{G3}.c_3, \, \mbox{for other,}\\
  m(\Theta)=1-(m(A)+m(B)+m(C)+m(D)+m(E)+m(F)+m(G)),\\
  \end{array}
  \right. \\
\end{array}
\end{eqnarray}
where $c_1$, $c_2$ and $c_3$ are the weights associated to the certitude respectively: ``sure'', ``moderately sure'' and ``not sure''. The chosen weights are here: $c_1=2/3$, $c_2=1/2$ and $c_3=1/3$. Indeed we have to consider the cases when the same kind of sediment (but with different certainties) is present on the same tile. The proportion of each sediment in the tile associated to these weights is noted, for instance for $A$: $p_{A1}$, $p_{A2}$ and $p_{A3}$.

The total conflict between the three experts is 0.2244. This conflict comes essentially from the difference of opinion of the experts and not from the tiles with more than one sediment. Indeed, we have a weak {\it auto-conflict} (conflict coming from the combination of the same expert three times). The values of the auto-conflict for the three experts are: 0.0496, 0.0474, and 0.0414. We note a difference of decision between the three combination rules giving by the equations (\ref{GeneDSmTcombination}) for the PCR6, (\ref{DP}) for the mixed rule and (\ref{conjunctive}) for the conjunctive rule. The proportion of tiles with a different decision is 0.11\% between the mixed rule and the conjunctive rule, 0.66\% between the PCR6 and the mixed rule, and 0.73\% between the PCR6 and the conjunctive rule.

These results show that there is a difference of decision according to the combination rules with the same model. However, we can not know what is the best decision, and so what is the best rule, because on this application no ground truth is known. We compare these same rules in another application, where the reality is completely known.

\section{Classifiers fusion in Radar target recognition}
\label{ranar}
Several types of classifiers have been developed in order to extract the information for the automatic target recognition (ATR). We have noted that these performances are different according to the classifier and the radar target. We have proposed different approaches of information fusion in order to outperform three radar target classifiers \cite{Martin04}. We present here the results reached by the fusion of three classifiers with the conjunctive rule, the DSmH, the PCR5 and the PCR6.

\subsection{Classifiers}
The three classifiers used here are the same than in \cite{Martin04}. The first one is a fuzzy $K$-nearest neighbor classifier, the second one is a multilayer perceptron (MLP) that is a feed forward fully connected neural network. And the third one is the SART (Supervised ART) classifier \cite{Radoi99} that uses the principle of prototype generation like the ART neural network, but unlike this one, the prototypes are generated in a supervised manner.

\subsection{Database}
\begin{figure}[htb]
\begin{center}
\includegraphics[height=5cm]{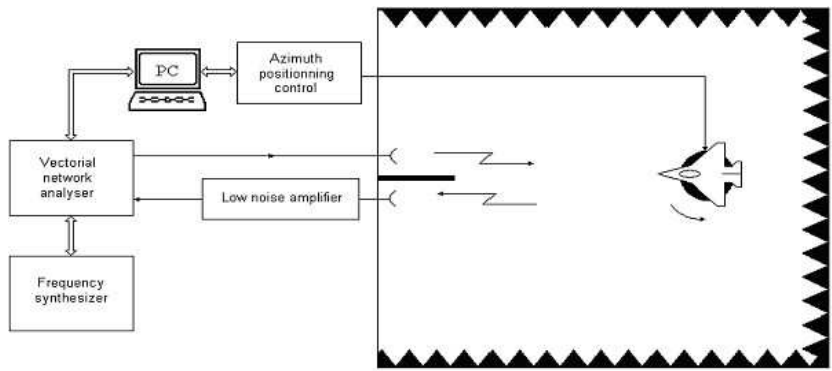}
\end{center}
\caption{Experimental setup.}
\label{chamber}
\end{figure}

The database is the same than in \cite{Martin04}. The real data were obtained in the anechoic chamber of ENSIETA (Brest, France) using the experimental setup shown on figure \ref{chamber}. We have considered 10 scale reduced (1:48) targets (Mirage, F14, Rafale, Tornado, Harrier, Apache, DC3, F16, Jaguar and F117). 

Each target is illuminated in the acquisition phase with a frequency stepped signal. The data snapshot contains 32 frequency steps, uniformly distributed over the band $B=[11650,17850]$MHz, which results in a frequency increment of $\Delta f=200$MHz. Consequently, the slant range resolution and ambiguity window are given by:
\begin{eqnarray}
\Delta R_s=c/(2B)\simeq 2.4m, \, \, W_s=c/(2\Delta f)=0.75m.
\end{eqnarray}

The complex signature obtained from a backscattered snapshot is coherently integrated via FFT in order to achieve the slant range profile corresponding to a given aspect of a given target. For each of the 10 targets 150 range profiles are thus generated corresponding to 150 angular positions, from -50 degrees to 69.50 degrees, with an angular increment of 0.50 degrees.

The database is randomly divided in a training set (for the three supervised classifiers) and test set (for the evaluation). When all the range profiles are available, the training set is formed by randomly selecting 2/3 of them, the others being considered as the test set.

\subsection{Model}
The numerical outputs of the classifiers for each target and each classifier, normalized between 0 and 1, define the masses. In order to keep only the most credible classes we consider the two highest values of these outputs referred as $o_{ij}$ for the $j^{th}$ classifier and the target $i$. Hence, we obtain only three focal elements (two targets and the ignorance $\Theta$). 

The classifier does not provide equivalent belief in mean. For example, the fuzzy $K$-nearest neighbors classifier provide easily a belief of 1 for a target, whereas the two other classifiers provide always belief not null on the second target and ignorance. In order to give the same weight to each classifier, we weight each belief by an adaptive threshold given by:
\begin{eqnarray}
f_j=\frac{0.8}{mean(o_{ij})}.\frac{0.8}{mean(b_{ij})},
\end{eqnarray}
where $mean(o_{ij})$ is the mean of the belief of the two targets on all the previous considered signals for the classifier $j$, $mean(b_{ij})$ is the similar mean on $b_{ij}=f_j.o_{ij}$. $f_j$ is initialized to 1. Hence, we expect the mean of belief on the targets tends toward 0.8 for each classifier, and 0.2 on $\Theta$. 

Moreover, if the belief mass on $\Theta$ for a given signal and classifier is less than 0.001, we keep the maximum of the mass and force the other in order to reach 0.001 on the ignorance and so avoid total conflict with the conjunctive rule.

\subsection{Results}

We have conducted the division of the database into training database and test database, 800 times in order to estimate better the good-classification rates. We have obtained a total conflict of 0.4176. The auto-conflict, reached by the combination of the same classifier three times, is 0.1570 for the fuzzy $K$-nearest neighbor, 0.4055 for the SART and 0.3613 for the multilayer perceptron. The auto-conflict for the fuzzy $K$-nearest neighbor is weak because it happens many times that the mass is only on one class (and ignorance), whereas there are two classes with a non-null mass for the SART and multilayer perceptron. Hence, the fuzzy $K$-nearest neighbor reduce the total conflict during the combination. The total conflict is here higher than in the previous application, but it comes here from the modelization essentially and not from a difference of opinion giving by the classifiers.

The proportion of targets with a different decision is giving in
percentage, in the table \ref{dis_rules}. These percentages are more
important for this application than the previous application on sonar
images. Hence the conjunctive rule and the mixed rule are very
similar. In terms of similarity, we can give this order: conjunctive
rule, the mixed rule (DP), PCR6f and PCR6g with a concave mapping,
PCR6, PCR6f and PCR6g with a convex mapping, and PCR5.

\begin{table}
\begin{center}
  \begin{tabular}{|l|c|c|c|c|c|c|c|c|}
    \hline
    Rule & \scriptsize Conj. & \scriptsize DP &
    \scriptsize PCR$f_{\sqrt{x}}$ & \scriptsize PCR$g_{\sqrt{x}}$ &
    \scriptsize PCR6 & \scriptsize PCR$g_{x^2}$ &  \scriptsize
    PCR$f_{x^2}$ & \scriptsize PCR5 \\
    \hline
    Conj. & 0 & 0.68 & 1.53 & 1.60 & 2.02 & 2.53 & 2.77 & 2.83 \\
    \hline
    DP & 0.68 & 0 & 0.94 & 1.04 & 1.47 & 2.01 & 2.27 & 2.37 \\
    \hline
    PCR$f_{\sqrt{x}}$ & 1.53 & 0.94 & 0 & 0.23 & 0.61 & 1.15 & 1.49 &
    1.67 \\
    \hline
    PCR$g_{\sqrt{x}}$ & 1.60 & 1.04 & 0.23 & 0 & 0.44 & 0.99 & 1.29 &
    1.46 \\
    \hline
    PCR6 & 2.04 & 1.47 & 0.61 & 0.44 & 0 & 0.55 & 0.88 & 1.08 \\
    \hline
    PCR$g_{x^2}$ & 2.53 & 2.01 & 1.15 & 0.99 & 0.55 & 0 & 0.39 & 0.71 \\
    \hline
    PCR$f_{x^2}$ & 2.77 & 2.27 & 1.49 & 1.29 & 0.88 & 0.39 & 0 & 0.51\\
    \hline
    PCR5 & 2.83 & 2.37 & 1.67 & 1.46 & 1.08 & 0.71 & 0.51 & 0 \\
    \hline
  \end{tabular}
\end{center}
\caption{Proportion of targets with a different decision (\%)\label{dis_rules}}
\end{table}

The final decision is taken with the maximum of the pignistic
probabilities. Hence, the results reached by the generalized PCR are
significantly better than the conjunctive rule and the PCR5, and
better than the mixed rule (DP). The conjunctive rule and the PCR5
give the worth classification rates on these data (there is no
significantly difference), whereas they have a high proportion of
targets with a different decision.

The best classification rate (see table \ref{good_classif}) is
obtained with PCR$f_{\sqrt{x}}$, but is not significantly better than
the results obtained with the other versions PCR$f$, using a different
concave mapping.

\begin{table}
\begin{center}
  \begin{tabular}{|l|c|c|}
    \hline
    Rule & \% & confiance Interval \\
    \hline
    Conjunctive & 89.83 & [89.75 : 89.91]  \\
    \hline
    DP & 89.99 & [89.90 : 90.08] \\
    \hline
    PCR$f_{x^{0.3}}$ & 90.100 & [90.001 : 90.200] \\
    \hline
    PCR$f_{\sqrt{x}}$ & {\bf 90.114} & [90.015 : 90.213] \\
    \hline
    PCR$f_{x^{0.7}}$ & 90.105 & [90.006 : 90.204] \\
    \hline
    PCR$g_{\sqrt{x}}$ & 90.08 & [89.98 : 90.18]\\
    \hline
    PCR6 & 90.05 & [89.97 : 90.13]  \\
    \hline
    PCR$g_{x^2}$ & 90.00 & [89.91 : 90.10]\\
    \hline
    PCR$f_{x^2}$ & 89.94 & [89.83 : 90.04] \\
    \hline
    PCR5 & 89.85 & [89.75 : 89.85]  \\
    \hline
  \end{tabular}
\end{center}
\caption{Good-classification rates (\%)\label{good_classif}}
\end{table}

\section{Conclusion}

In this chapter, we have proposed a study of the combination rules compared in terms of decision. The generalized proportional conflict redistribution (PCR6) rule (presented in the chapter \cite{Martin06a}) have been evaluated. We have shown on real data that there is a difference of decision following the choice of the combination rule. This difference can be very small in percentage but allows significantly difference in good-classification rates. Moreover, high proportion with a different decision does not lead to a high difference in terms of good-classification rates. The last application shows that we can achieve better good-classification rates with the generalized PCR6 than with the conjunctive rule, the DSmH, or PCR5.

The first presented application shows that the modelization on $D^\Theta$ can resolve easily some problems. If the application need a decision step and if we want to consider the conjunctions of the elements of the discernment space, we have to take the decision directly on the masses (and not on the credibilities, plausibilities or pignistic probabilities). Indeed, these functions are increasing and can not give a decision on the conjunctions of elements. In real applications, most of the time, there is no ambiguity and we can take the decision, else we have to propose a new decision function that can reach a decision on conjunctions and also on singletons.

The conjunctions of elements can be considered (and so $D^\Theta$) in many applications, especially in image processing, where an expert can provide element with more than one classes. In estimation applications, where intervals are considered, encroaching intervals (with no empty intersection) can provide better modelization.

\bibliographystyle{plain} 
\bibliography{biblio}

\end{document}